\documentclass[journal]{vgtc}                

\usepackage[export]{adjustbox}
\usepackage{appendix}
\usepackage{array}
\usepackage{enumitem}
\usepackage{graphicx}
\usepackage{tikz}
\usepackage{times}
\usepackage{hyperref}
\usepackage{amsmath}
\usepackage{amssymb}
\usepackage{float}
\usepackage{adjustbox}
\usepackage{booktabs}
\usepackage{wrapfig}
\usepackage{stfloats}
\usepackage{subcaption}

\usepackage[none]{hyphenat}
\emergencystretch=\maxdimen
\onlineid{0}

\vgtccategory{Research}

\vgtcinsertpkg

\title{Exploring the Effectiveness of Dataset Synthesis: \\ An application of Apple Detection in Orchards \\ \normalsize{NHL Stenden Professorship in Computer Vision \& Data Science}} 

\author{Alexander van Meekeren \\ \small{Supervisors: Maya Aghaei Gavari, Klaas Dijkstra}}
\authorfooter{
\item
  Alexander van Meekeren is an Applied Chemistry student at the NHL Stenden University of Applied Sciences, E-mail: alexander.van.meekeren@student.nhlstenden.com.
\item
  Maya Aghaei Gavari is a researcher at the NHL Stenden Professorship in Computer Vision \& Data Science, E-mail: maya.aghaei.gavari@nhlstenden.com.
\item
  Klaas Dijkstra is a professor at the NHL Stenden Professorship in Computer Vision \& Data Science, E-mail: klaas.dijkstra@nhlstenden.com.
}

\abstract{Deep object detection models have achieved notable successes in recent years, but one major obstacle remains: the requirement for a large amount of training data. Obtaining such data is a tedious process and is mainly time consuming, leading to the exploration of new research avenues like synthetic data generation techniques. In this study, we explore the usability of Stable Diffusion 2.1-base for generating synthetic datasets of apple trees for object detection and compare it to a baseline model trained on real-world data. After creating a dataset of realistic apple trees with prompt engineering and utilizing a previously trained Stable Diffusion model, the custom dataset was annotated and evaluated by training a YOLOv5m object detection model to predict apples in a real-world apple detection dataset. YOLOv5m was chosen for its rapid inference time and minimal hardware demands. Results demonstrate that the model trained on generated data is slightly underperforming compared to a baseline model trained on real-world images when evaluated on a set of real-world images. However, these findings remain highly promising, as the average precision difference is only 0.09 and 0.06, respectively. Qualitative results indicate that the model can accurately predict the location of apples, except in cases of heavy shading. These findings illustrate the potential of synthetic data generation techniques as a viable alternative to the collection of extensive training data for object detection models.
	
	\begin{minipage}[t]{1\linewidth}
	\end{minipage}
} 


\begin{document}
\firstsection{Introduction}
\maketitle

Deep learning has revolutionized the field of computer vision, particularly in object detection algorithms. Numerous state-of-the-art applications rely on deep learning techniques to achieve remarkable performance \cite{drone1, drone2}. However, despite the widespread adoption of deep learning in computer vision, the availability of data for training these algorithms remains a significant challenge. Acquiring suitable image data is a time-consuming process that involves multiple stages, such as sourcing relevant data, manual data collection, and data cleaning to ensure its representativeness for the real-world problem. Moreover, it demands considerable human effort and meticulous attention to detail to ensure the quality of the collected data. This often entails the need to hire data collectors, establish robust data-gathering infrastructure, and maintain data quality over time to ensure accuracy, completeness, and reliability. The scarcity of readily available and high-quality big data poses a significant hurdle in the development and deployment of deep learning models for computer vision tasks.

To address this issue, the development of research to generate synthetic training data via image synthesis emerged. Training data generated with image synthesis, in this research, will be utilized to train object detectors, to be used by a UAV system for various applications. Examples are aerial photography \cite{drone_photo_1, drone_photo_2}, crop monitoring \cite{drone_crop_1, drone_crop_2}, and inspection of infrastructure \cite{drone_infra_1, drone_infra_2}. Using image synthesis, this method eliminates the need for vast amounts of real-world data to train deep learning models and allows for the creation of training sets that can simulate a variety of scenarios.

Image synthesis can be accomplished using a variety of techniques \cite{GAN, VAE, dalle}. However, current research indicates that Latent Diffusion models are a promising direction \cite{LatentDM}. One example of such models is Stable Diffusion, an image generation model that was introduced in 2022 \cite{LatentDM, What_is_TTI}. This model operates by converting user-provided text prompts, which are text-based keywords and descriptions, into corresponding images \cite{how_to_prompt}.

While data acquisition for training deep learning models is a general issue, the feasibility of this approach will focus on synthesizing hyperrealistic, also known as photorealism, images of apple orchards. Apples are relatively simple objects, making them ideal for experimentation. Simultaneously, there exist well-established benchmark datasets for this task \cite{apple_database}.  The aim of this research is to synthesize images of apple orchards using Stable Diffusion and annotating the location of the apples, ultimately for the purpose of training deep learning models, such as YOLOv5 \cite{yolov5}. With this in mind, the following questions have been formulated:

\begin{enumerate}
    \item How might Latent Diffusion be utilized to synthesize images of apple orchard, that resemble real-world like scenarios?
    \item What is the performance of a deep learning model trained on synthetic data versus a deep learning model trained on real data, when tested in a real-world scenario?
\end{enumerate}

It is hypothesized that deep learning models trained on synthetic data, that resemble real-world like images, exhibit comparable performance to models trained on real data when evaluated in real-world scenarios. This hypothesis suggests that synthetic data can effectively train models to achieve similar levels of performance as those trained on real data, indicating the potential of synthetic data as a viable alternative or supplement to real data in deep learning model training.

The proposed research aims to enhance the current data collection process by introducing an automated approach that uses artificially synthesized images to train deep learning models. This could lead to more efficient and cost-effective data collection and practical applications in various industries.

\section{State of the art}

Image synthesis and object detection have emerged as powerful tools with state-of-the-art performance in a diverse field of machine learning and deep learning applications. For this research, a few key concepts will be addressed.

\subsection{Object Detection}
Object detection is a computer vision technique that involves identifying and locating objects within digital images or video frames. This is accomplished by analyzing the content of a given image, looking for patterns or features that are unique to specific objects or object categories, and then determining the location of each object within the image \cite{object_detec_dl}.

There are several approaches for object detection; nevertheless, one of the most popular is the use of deep neural networks \cite{DNN}. These networks are often trained on large datasets of images and annotated with information, such as bounding boxes, which describe information about the objects contained within them. Object detection networks learn to recognize features common to specific objects or object categories and use these features to predict objects in new images \cite{object_detec_dl}.

YOLOv5, released in 2021 \cite{yolov5}, is an example of such popular object detection algorithm. It is an improvement on the YOLO (You Only Look Once) family of real-time object detection models \cite{yolov5_or_dox}. When trained, an input image is fed into the YOLOv5 network and processed through a deep convolutional neural network (CNN) to extract distinctive features, followed by a feature pyramid network to detect objects of various sizes. YOLOv5 predicts, among others, the class, the location and provides a confidence score for the object in question.

Other widely used object detection algorithms include Faster-RCNN (Region-based Convolutional Neural Networks) \cite{frcnn, RCNN} and RetinaNet \cite{Retina}. However, in terms of inference time, they tend to be slower compared to YOLOv5. The underlying reason for this disparity lies in the inherent slower and more computationally demanding nature of Faster-RCNN and RetinaNet networks themselves \cite{yolo_rcnn_comparison, yolo_retina_comparison}. Faster R-CNN's computationally expensive nature arises from its utilization of a region proposal network (RPN) \cite{frcnn} and a classification network. In contrast, the faster performance of YOLOv5 stems from its adoption of a single neural network for direct prediction. RetinaNet, a two-stage framework, incorporates a backbone network, feature pyramid network (FPN) \cite{fpn}, and task-specific subnetworks. YOLOv5, a one-stage method, prioritizes efficient inference.

Since the object detector will be used in UAVs, it is crucial that it can accurately predict objects in real-time. An important application of real-time object detection is environmental awareness, which greatly enhances a drone's situational awareness. By understanding the surrounding environment, drones can adjust their behavior accordingly \cite{drone_env}. Considering the need for fast data processing, YOLOv5 is the preferred choice due to its faster inference time. Hence, YOLOv5 is the selected approach for real-time object detection in UAVs and will be used in this research.

\subsection{Image Synthesis}

\begin{figure*}[ht]
    \centering
    \includegraphics[width=1\textwidth, height=0.2\textheight]{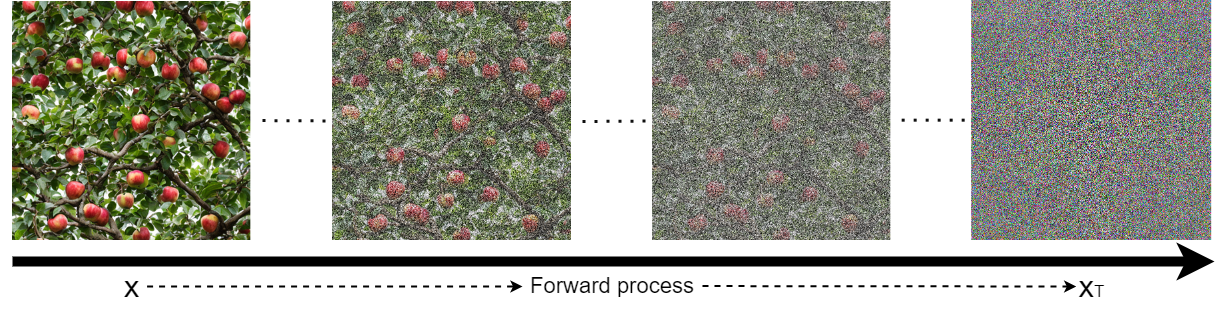}
    \caption{
        Schematic representation of the forward process. Gaussian noise gets added to the image $x$ in $t$ steps until the noised image $x_T$ is formed.}
    \label{fig:forward}
\end{figure*}

Image synthesis, in computer vision, is the computational process of producing previously unseen images through algorithmic manipulation via computer programs. This can involve generating images from scratch, modifying existing images, or combining multiple images to create new compositions \cite{KUMAR2020167}. Image generation can be accomplished using various techniques, including deep learning models such as generative adversarial networks (GANs) \cite{GAN_publication} and variational autoencoders (VAEs) \cite{ganvae, GAN_surv}. GANs use a generator and discriminator to compete and produce diverse samples of images. VAEs use an encoder and decoder to learn a latent space and this to generate samples of images with better control. 

Another deep learning approach to image synthesis that has gained popularity recently are diffusion models \cite{deep_uns_learn}. By incrementally adding Gaussian noise to an image and then gradually diffusing that noise, diffusion models are able to synthesize realistic images. An example is the diffusion model developed by Ho et al. \cite{diffmodels, diff_beat_gan, improved_dm}. This model works on the same principle and has produced high-quality images with encouraging results.

Although these image synthesis techniques with utilizing solely diffusion models are impressive in their own right, recent years have witnessed the emergence of additional approaches, including a conditional image synthesis technique discussed in the next section.

\subsection{Conditional Image Synthesis}

Text-to-image is a technique that uses deep learning models to generate realistic images from textual descriptions. This is accomplished by conditioning the generation process of these images on text-based prompts. In text-to-image diffusion models, the input text is usually first processed by a Transformer-based model \cite{aiayn}, which generates a series of text embeddings from the text. A noisy image is fed into a series of convolutional layers and gets ultimately merged with the text embedding information, via a cross-attention module \cite{aiayn}, which incrementally generates the image by refining and adding details to the initial image representation. \cite{con_t2i, gan_t2i}.

DALL-E 2 is an example of the text-to-image approach and is developed by OpenAI \cite{dalle2}.  It builds on the success of OpenAI's original DALL-E model, which was introduced in 2021 \cite{dalle}. The original DALL-E model could generate novel images based on textual input, such as “Astronaut riding a horse”.  DALL-E 2 takes this approach further by generating even more complex and detailed images based on more complex textual inputs. One issue remains, The closed-source status of DALL-E 2 presents a limitation for its utilization in specific projects, as access to the underlying source code is restricted for the general public.

Another text-to-image generative model that is open source and based on diffusion models is Latent Diffusion.  Latent Diffusion reduces the computational demand of diffusion models and text-to-image techniques by converting images into latent representations using a VAE \cite{LatentDM}. These latent representations are used to speed up image synthesis and offer a more effective and efficient representation than raw image space \cite{LatentDM, weng2021diffusion}. By performing cross attention \cite{aiayn} on these latent representations together with the text embeddings, the generated images  are coherent with the given prompts. 

A notable example of such models is Stable Diffusion, an image generation model that was introduced in 2022 \cite{LatentDM, What_is_TTI}. This model operates by converting prompts, into corresponding images \cite{how_to_prompt}. The model was trained on large datasets of images and corresponding text-based prompts to generate high-quality images that are consistent with given prompts.

\subsection{Prompt Engineering}

A useful technique for obtaining realistic images in Stable Diffusion involves utilizing Prompt Engineering \cite{prompt_engineering}. This entails creating input prompts that can control the image generation process, including initial images or masks, as well as textual or other cues that can be fed into the model as input.

Prompt engineering enables the creation of highly specific and detailled images using Stable Diffusion. It can be employed to generate images of particular objects, scenes, or styles, as well as to manipulate the image generation process to achieve desired artistic effects.

To generate high-quality images using Stable Diffusion, carefully crafted positive and negative prompts can be employed. The inclusion of negative prompts in Stable Diffusion \cite{github-stable-diffusion} was a refinement over Latent Diffusion \cite{LatentDM} and involved examining the distinction between the image that is being generated, to steer the final image towards the positive prompt and steer away from the negative prompt. 

While prompt-based image generation can produce realistic images, using Classifier-free Guidance (CFG) \cite{cfg} can provide even greater control over the generation process. CFG, in Stable Diffusion, amplifies the effect of the text prompt on the generated image. By default, Stable Diffusion applies a classifier to the text prompt to guide the generation of the image \cite{LatentDM}. However, CFG allows for fine-tuning the influence of this classifier, resulting in more creative control over the generated image. Typically, CFG is defined to be in a range between 1 and 30, with lower values generating more creative images.

More specifically, CFG determines the trade-off between the coverage of modes and image fidelity \cite{cfg}. When set to 1, the model generates samples based solely on the prior distribution, without any guidance. As the guidance scale increases, the model is instructed to produce samples that better match some given condition. The classifier-free aspect of this technique refers to the fact that it does not require training a classifier to incorporate guidance during generation \cite{cfg}. Instead, the model is guided by adding a penalty term to the generation process, forcing the generated images to match the desired condition.

\section{Methodology}

\begin{figure}[H]
    \centering
    \includegraphics[width=0.455\textwidth, height=0.11\textheight]{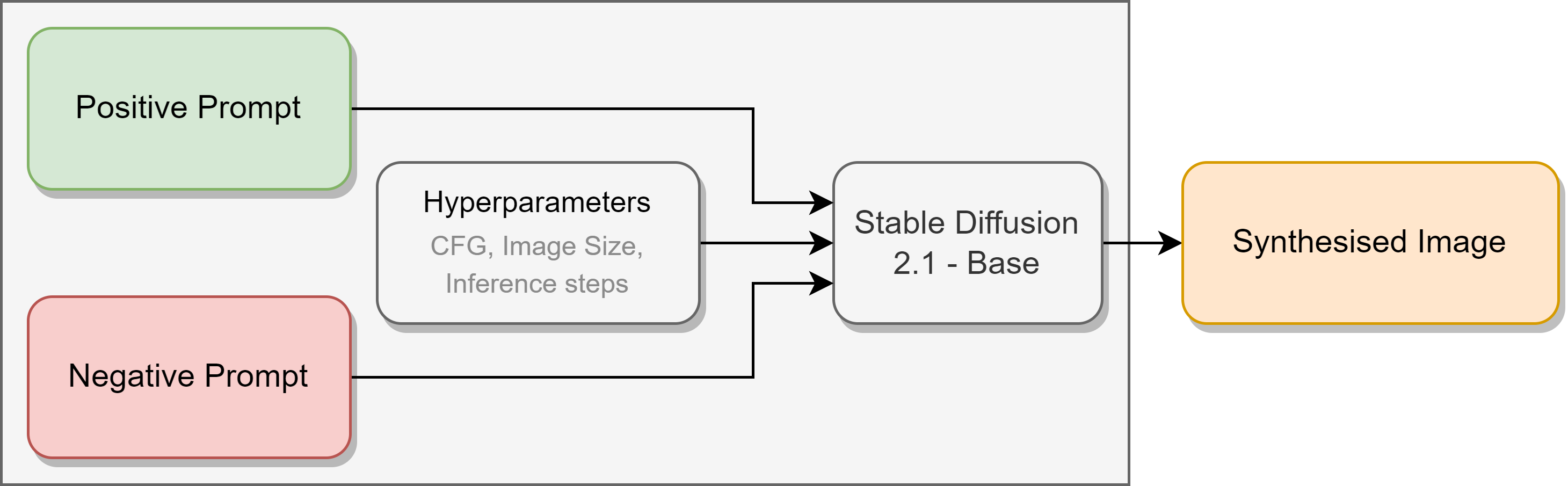}
    \caption{
        Schematic representation of the image synthesis process.}
    \label{fig:mm}
\end{figure}

In this study, a YOLOv5m model was trained on a custom dataset of apple trees generated using Stable Diffusion 2.1 - base, primarily with the help of positive and negative prompts. An overview of the image generation process can be seen in Figure \ref{fig:mm}. The custom dataset was then annotated with auto annotation and bounding box filtering techniques. Subsequently, the YOLOv5 model was trained on the annotated dataset of apple trees. To assess the effectiveness of the model trained on synthesized images, a second YOLOv5 model was trained on the pre-existing MinneApple dataset \cite{apple_database} as the baseline. The performance of the two models was determined by comparing their average precision on the MinneApple testing set.

\subsection{MinneApple dataset}

\begin{figure}[ht]
    \centering
    \includegraphics[width=0.4\textwidth, height=0.24\textheight]{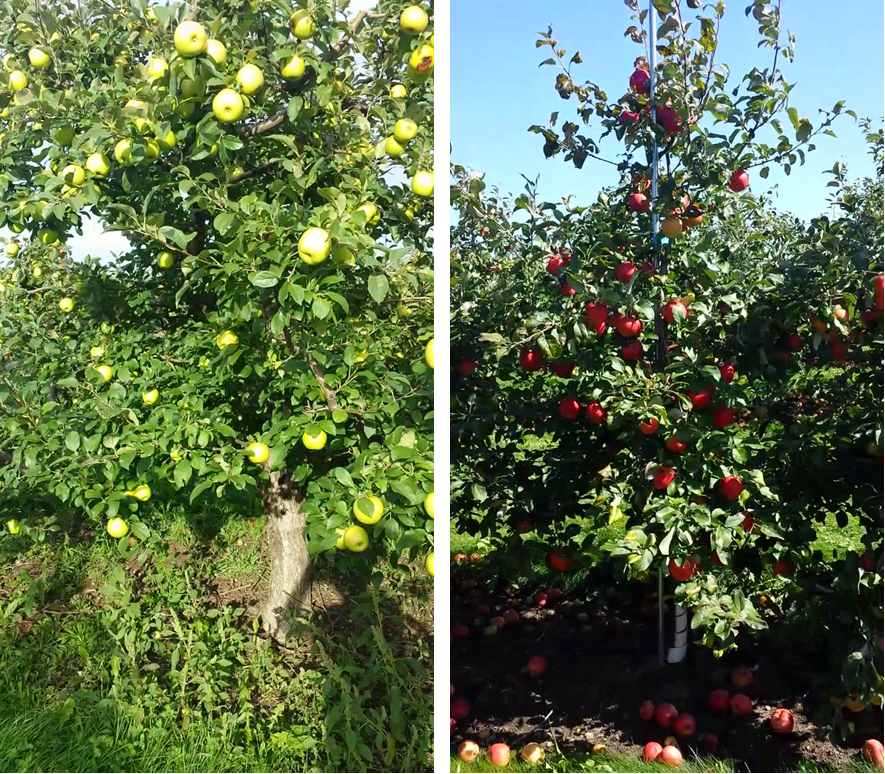}
    \caption{
        Sample images from the MinneApple training dataset}
    \label{fig:minneapple}
\end{figure}

The MinneApple dataset is a benchmark dataset for apple detection and segmentation \cite{apple_database, minneapple2}. It contains data for patch-based counting of clustered fruits and is labeled using polygonal masks for each object instance to aid in precise object detection, localization, and segmentation. The size and diversity of the MinneApple dataset are two of its key strengths. The dataset includes 1001 images in total, with 670 images for training and 331 images for testing, and has a standardized resolution of 1280x720 pixels. Each image was taken from a different angle and under different lighting conditions, as can be seen in Figure \ref{fig:minneapple}. As a result, the dataset covers a wide range of scenarios, making it an excellent benchmark for testing and evaluating various apple detection and segmentation models. 

For this reason, the MinneApple dataset will serve as the baseline for evaluating object detection models \cite{apple_database, minneapple2}. The MinneApple dataset will be utilized for training and testing the baseline model. Additionally, the MinneApple data will be used to test the model trained on generated data. It is important to note that the MinneApple dataset does not contain a validation set. Therefore, the MinneApple training dataset will be divided into two distinct subsets to facilitate training: a training set and a validation set. These subsets will adhere to an 80:20 ratio, with the training set comprising 536 images and the validation set containing 134 images.

\subsection{Diffusion Models}

\begin{figure*}[ht]
    \centering
    \includegraphics[width=1\textwidth, height=0.25\textheight]{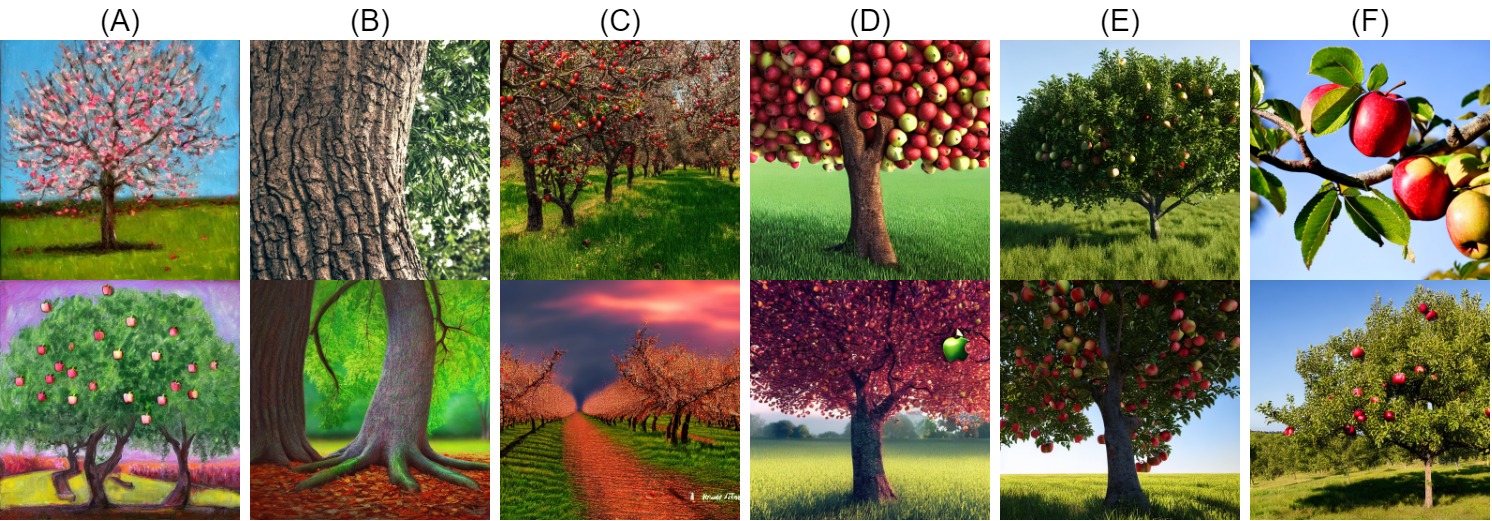}
    \caption{
        Synthesized images using Stable Diffusion, with the default model configurations. See Appendix \ref{app:synthesize-images} for additional examples. A) Prompt: \textit{"Apple trees"}. B) Prompt: \textit{“photo of a tree, hyperrealism, 4k, realistic, photograph"}. C) Prompt: \textit{“apple orchard, hyperrealism, 4k, realistic, photograph"}. D) Prompt: \textit{“apple tree with many apples, apples, hyperrealism, 4k, render, cinematic lighting"}. E) Positive prompt: \textit{prompt D)}; Negative prompt: \textit{“blurry image, deformed, cartoon, drawing"}. F) Positive prompt: \textit{“photo of a tree branch with apples, apple tree, many apples, hyperrealism, 4k, realistic, photograph"}; Negative prompt: \textit{negative prompt E)}.
        }
    \label{fig:apple_1-mini}
\end{figure*}

Diffusion models (DM) are a type of probabilistic model that can effectively model any data distribution by employing both a so-called forward and backward process \cite{deep_uns_learn}. In the context of image synthesis, the forward process involves adding Gaussian noise to a given image at $t$ intervals. A Gaussian distribution is preferred as it is simple to sample from. The backward process, on the other hand, aims to learn how to reverse the original noising process by predicting the noise that was added to the image. This process is typically implemented using an UNet network \cite{Unet} that acts as a series of denoising autoencoders \cite{diff_beat_gan}.

More precisely, the forward process can be viewed as a Markov Chain of length $T$, where at each time step $t$, a new image $x_t$ is generated by adding Gaussian noise to the previous image $x_{t-1}$, see Equation \ref{eq:xt}. This process is illustrated in Figure \ref{fig:forward} \cite{deep_uns_learn}.

\begin{equation}
    \label{eq:xt}
    \mathnormal{x_t := x_{t-1} + \epsilon_t}
\end{equation}

$\epsilon_t$ is the added Gaussian noise at time step $t$. Thus, the state of the system at time step $t$ is given by the image $x_t$, which is a function of the previous state $x_{t-1}$ and the current noise $\epsilon_t$ \cite{deep_uns_learn}. The backward process involves an autoencoder, typically implemented using an UNet architecture \cite{Unet}, that tries to predict the noise $\epsilon_t$, given the image $x_t$ and time step $t$ \cite{deep_uns_learn}. This can be seen in Equation \ref{eq:ept}. 

\begin{equation}
    \label{eq:ept}
     \epsilon_\theta(\mathnormal{x}_t, t) \approx \epsilon_t
\end{equation}

Where $\epsilon_\theta$ is the function that predicts the noise $\epsilon_t$ added to the image $x_t$. Therefore, the state transition in the backward process involves predicting the noise at each time step, which is used to reconstruct the previous image \cite{deep_uns_learn, LatentDM}.

The backward process is responsible for predicting noise, which can then be used to reconstruct the original image. To accomplish this, the loss function is defined as the predicted noise's Mean Square Error (MSE). In this context, the MSE function is a reconstruction loss and will be minimized during training. Equation \ref{eq:DM} represents the mathematical expression for the loss function \cite{LatentDM}.

\begin{equation}
    \label{eq:DM}
    L_{DM}:=\mathbb{E}_{x,\epsilon \sim \mathnormal{N}(0, I), t} \Bigr{[}\|\epsilon_t-\epsilon_\theta(\mathnormal{x}_t, t)\|_2^2\Bigr{]}
\end{equation}

The traditional method of generating images through diffusion models suffers from a slow sampling time due to sequential evaluations. Latent Diffusion models (LDMs) have been proposed as a solution to this problem. To improve computational efficiency, LDMs use lower dimensional latent representations of images learned using a variational autoencoder (VAE) \cite{VAE}. To accomplish this, the model's loss function is modified to account for the change in representation, as expressed by Equation \ref{eq:LDM}. $\mathcal{E}(x)$ represents the encode function of the VAE and $\mathnormal{z}_t$, originating from $\mathcal{E}(x)$, is the latent representation of $x_t$ \cite{LatentDM}.

\begin{equation}
    \label{eq:LDM}
    L_{LDM}:=\mathbb{E}_{\mathcal{E}(\mathnormal{x}),\epsilon \sim \mathnormal{N}(0, I), t} \Bigr{[}\|\epsilon-\epsilon_\theta(\mathnormal{z}_t, t)\|_2^2\Bigr{]}
\end{equation}

\begin{figure}[H]
    \centering
    \includegraphics[width=0.48\textwidth, height=0.16\textheight]{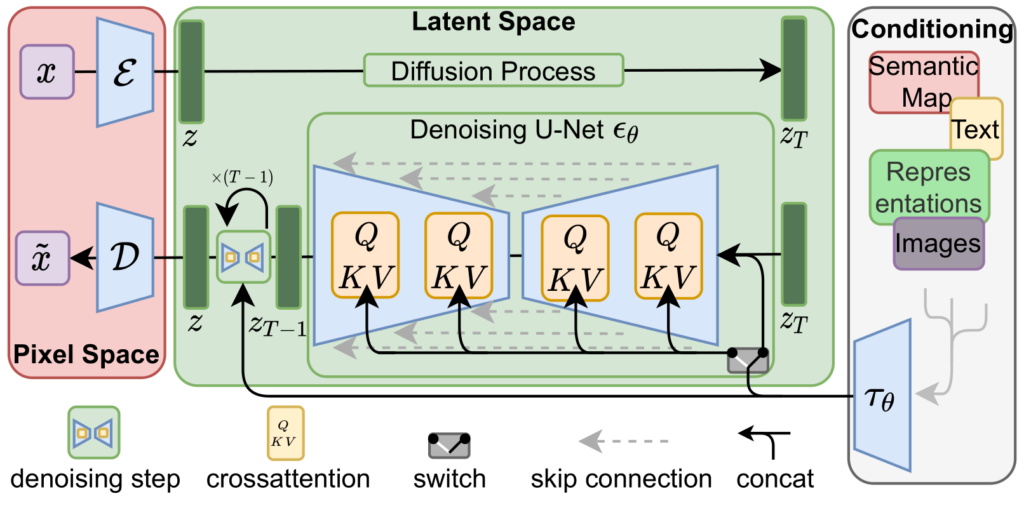}
    \caption{
        Schematic representation of the forwards and backwards process \cite{LatentDM}. Including cross attention with the semantic compressions of the text prompts \cite{aiayn}.}
    \label{fig:pti}
\end{figure}

To condition the image generation process on prompts during the diffusion process, the prompts undergo transformation into semantic compressions using a text transformer $\mathnormal{T}_\theta$ such as CLIP \cite{CLIP}. These semantic compressions are subsequently concatenated with the most recent representation of the noisy latent image during the backward process. An illustration of this process is depicted in Figure \ref{fig:pti}. In particular, they are added to the denoising U-net through cross attention \cite{aiayn}, enabling the denoising process to be conditioned on the provided prompts. The outcome is a denoised image that represents the given prompt.

\subsection{Training Data Generation}

Images and their annotations are required to train object detectors. To collect this information, a method that synthesises and annotated images automatically was created. 

\subsubsection{Image Synthesis: Prompt Engineering}
\label{sec: iamge synthesis}

Images that resemble the MinneApple dataset were generated using the Stable Diffusion pipeline \cite{hf_diffusers_api}. The image synthesis process began  with the utilization of Runwayml's stable-diffusion-v1-5 model \cite{sd15_url}, which inherited the weights from Stable-Diffusion-v1-2. This model underwent fine-tuning by Runwayml using images from the laion-aesthetics v2 5+ dataset \cite{laion_v2}, containing images with a resolution 512x512 pixels. The exploration for effective prompts to generate real-lifelike apple tree images was initiated with the prompt "Apple trees." During this stage of the research, the pipeline adhered to standard hyperparameters, including a classifier-free guidance scale (CFG) \cite{cfg} of 7.5, an image output size of 512x512, and 50 inference steps. The default noise scheduler of the model, the Discrete Euler scheduler \cite{EulerDiscreteScheduler}, was used. The resulting images, as shown in Figure \ref{fig:apple_1-mini}A, exhibited drawings or paintings resembling apple trees, generic trees, and pink blossom trees. Thus, these images do not resemble hyperrealistic apple trees. 

The prompt was changed to synthesize the images, to \textit{“photo of a tree, hyperrealism, 4k, realistic, photograph”} to create more lifelike apple trees that closely resemble real-life trees. Although the model generated more realistic images of trees, as demonstrated in Figure \ref{fig:apple_1-mini}B, no apples were present in the generated images. In order to achieve more realistic images, the prompt was adjusted to generate apple orchards by incorporating the phrase: \textit{“apple orchard, hyperrealism, 4k, realistic, photograph”}. The resulting images, as shown in Figure \ref{fig:apple_1-mini}C, included apple trees with occasional apple appearances. However, \textit{if} apples were produced by the model that were in apple trees, they are frequently small and difficult to identify in most cases. To address this problem, a new prompt was created that highlighted the presence of apples in the generated images. This prompt, \textit{"apple tree with many apples, apples, hyperrealism, 4k, render, cinematic lighting"} allowed the model to concentrate more on the appearance of the apples. The outcome was the creation of semi-realistic apple trees with more recognizable apples, as illustrated in Figure \ref{fig:apple_1-mini}D.

Despite the previous improvements, the synthesized images still contained unrealistic or deformed apples. To further enhance the images and remove misshapen apples, negative prompts were introduced: \textit{“blurry image, deformed, cartoon, drawing”}. By implementing negative prompts, the realism of the images was improved, and this change effectively eliminated the appearance of misshapen apples in the synthesized images, yielding even more real-lifelike results, as shown in Figure \ref{fig:apple_1-mini}E. To further heighten the realism of the synthesized images, the prompt was modified to \textit{“photo of a tree branch with apples, apple tree, many apples, hyperrealism, 4k, realistic, photograph”} This adjustment emphasized the presence of apples even more, resulting in even more realistic images, as illustrated in Figure \ref{fig:apple_1-mini}F.

In order to make a fair comparison with the MinneApple dataset, the Stable Diffusion pipeline was modified to generate images matching the image size of MinneApple, resulting in images of 1280x704 pixels. Although not an exact match, the size was deemed close enough to be used for training purposes. The original images of the MinneApple dataset are incompatible with the custom YOLOv5 training pipeline, as YOLOv5 requires the image resolution to be dividable by a fixed power of two.

One of the challenges with the modified image size is due to its relatively large dimensions; requiring a significant amount of GPU memory, in combination with the current model being employed. Therefore, it was a necessity to switch to a different Stable Diffusion model, to alleviate the VRAM usage problem. To address this issue, a newer and smaller model called stable-diffusion-2-1-base from Stabilityai \cite{stable_diff21, sd21_url} was used in conjunction with minor optimizations, such as the utilization of memory efficient attention from xFormers \cite{xFormers}, to be able to generate images of the desired size. For this reason, this model was used for further image synthesis. Stable-diffusion -2-1-base is a fine-tuned version of the stable-diffusion-2-base trained on the Laion-5B dataset \cite{LAION}. The model was trained on images with 512 x 512 pixels. 

\begin{figure}[H]
    \centering
    \includegraphics[width=0.4\textwidth, height=0.25\textheight]{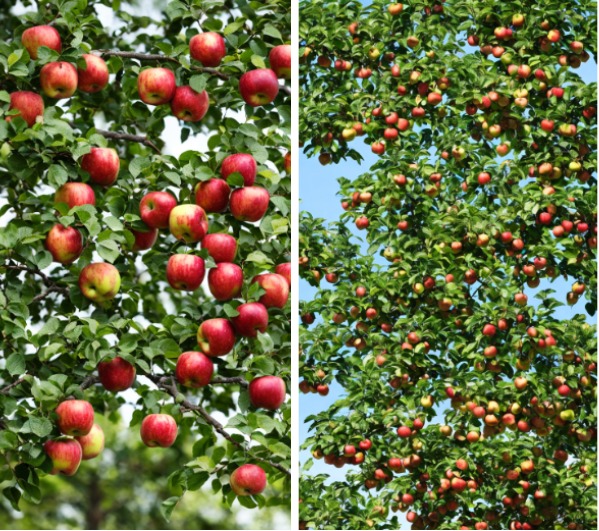}
    \caption{
        Synthesized images using Stable Diffusion, with the default model configurations. See Appendix \ref{app:synthesize-images} for additional examples. Both images are created with the same image size (1280x704) and prompts; Positive prompt: \textit{"photo of a tree branch with apples, apple tree, many apples, hyper-realism, 4k, realistic, photograph"}, Negative prompt: \textit{"blurry image, deformed, cartoon, drawing"}. (\textit{right}) has in addition the following modified parameters; CFG: 6, inference steps: 30.}
    \label{fig:apple_2-mini}
\end{figure}

The recently implemented model and new image size were used to generate images that closely resembled apple trees, as depicted in Figure \ref{fig:apple_2-mini} (\textit{left}). To optimize image synthesis, the class-free guidance was reduced to 6. With this reduction, the images resembled more trees than zoomed-in crops of apple tree foliage, as seen in Figure \ref{fig:apple_2-mini}. To reduce the generation time, the number of inference steps was decreased to 30 instead of 50. Despite this reduction, enough details were generated as shown in Figure \ref{fig:apple_2-mini} (\textit{right}). These adjustments resulted in highly realistic apple trees.

\begin{figure}[H]
    \centering
    \includegraphics[width=0.4\textwidth, height=0.25\textheight]{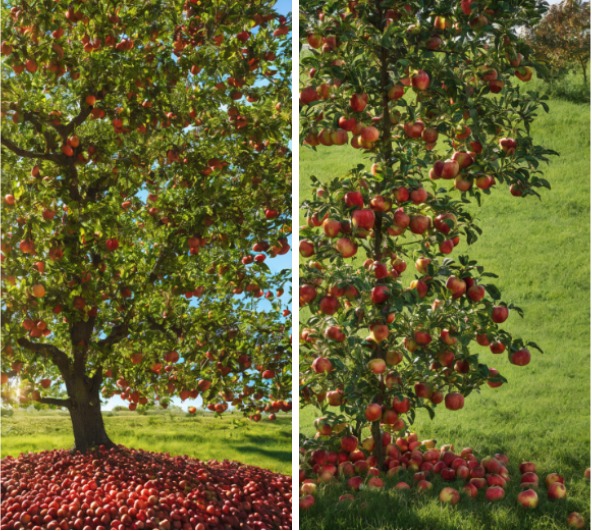}
    \caption{
        Synthesized images using Stable Diffusion, with the default model configurations. See Appendix \ref{app:synthesize-images} for additional examples. Positive prompt: \textit{"a photo of a tree standing in the grass. the tree has many apples, the apples are both red and yellow. beneath the tree there are a lot of apples. The many apples are a combination of red apples and yellow apples. volumetric lighting. shadows, hyperrealism, 4k realism, photograph"}, Negative prompt: \textit{"blurry image, deformed, cartoon, drawing, painting"}, image size: 1280x704. CFG: 6, inference steps: 30.}
    \label{fig:apple_final}
\end{figure}

Although the current prompting produces photo-like apple tree images,  the current images do not resemble the MinneApple dataset. The MinneApple data set contains entire apple trees, consisting of red and yellow apples, rather than a cropped image of the leaves with apples, as can be viewed in Figure \ref{fig:apple_2-mini}.  In addition, the MinneApple dataset contains unannotated apples on the ground near the trunk of the apple tree. A prompt has been created to generate images of full apple trees with apples on the ground, with an emphasis on red and yellow apples. Examples can be seen in Figure \ref{fig:apple_final}. This prompt replaces the previous prompt: \textit{"a photo of a tree standing in the grass. the tree has many apples, the apples are both red and yellow. beneath the tree there are a lot of apples. The many apples are a combination of red apples and yellow apples. volumetric lighting. shadows, hyperrealism, 4k realism, photograph"}

Based on the striking realism of the synthesized apple tree images depicted in Figure \ref{fig:apple_final} with the corresponding prompts, a new dataset was made for training the YOLOv5 model, consisting of 670 images with an 80:20 ration training validation split \cite{split_ratio}, both to ensure consistency with the MinneApple dataset.


\subsubsection{Automatic Annotation}


After the collection of the generated images, annotations of the apples in the generated data are needed. To do so, the manual annotation process was aided by using a pretrained detector to predict the bounding boxes in the generated images. 

Firstly, a Mask-RCNN network with a ResNet-50-FPN backbone \cite{mrcnn} was used to acquire the coordinates of the apples from the synthesized images.  This network was pre-trained on the MS-COCO dataset \cite{mscoco}, which contains images of apples.

By obtaining the coordinates of the apples for each image, the coordinates were transformed into bounding boxes. The bounding boxes were first filtered by the class label. All non-Apple classes were eliminated. After that, the Mask-RCNN confidence score was used to narrow down the remaining bounding boxes. Where the confidence threshold was set at 70\%. Non-max suppression was used for the final filter, with an IoU threshold of 0.2. This resulted in fewer redundant and double-bounding boxes for the apples, with only one bounding box for each apple. 

\begin{figure}[h]
    \centering
    \includegraphics[width=0.4\textwidth, height=0.25\textheight]{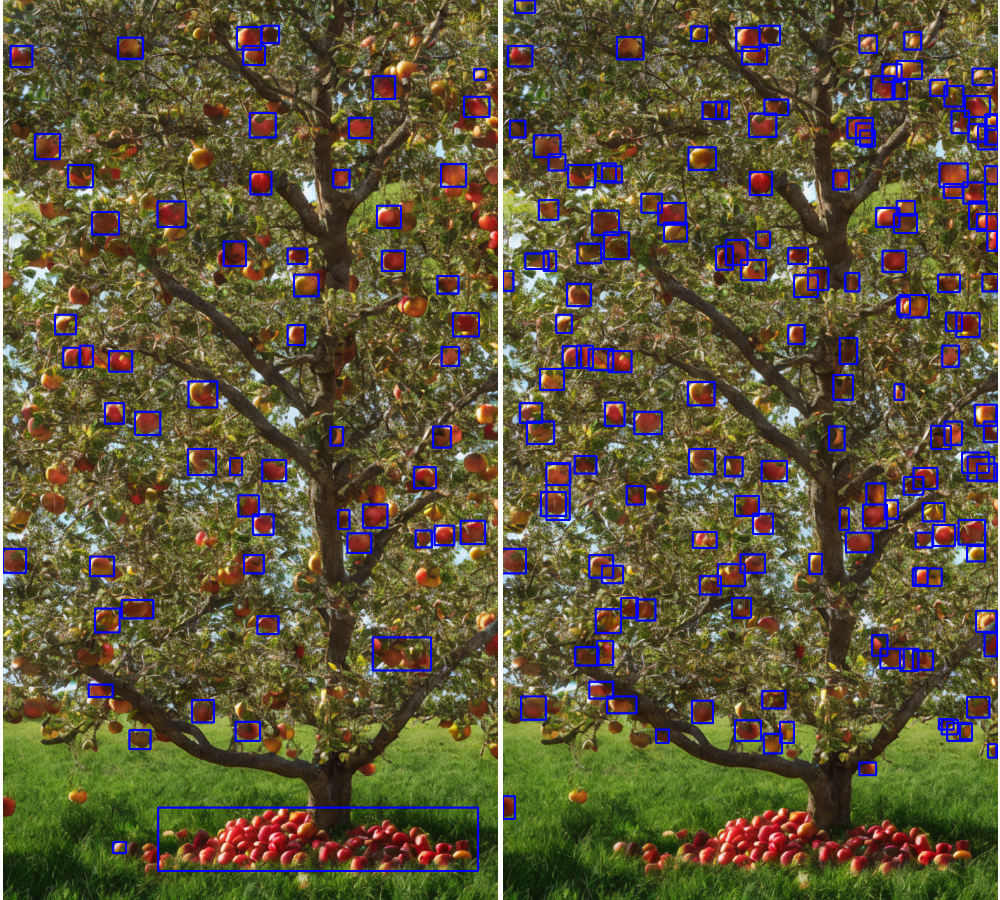}
    \caption{
        A sample image of the generated dataset, where (\textit{left}) the apples are annotated with the Mask-RCNN and (\textit{right}) the apples are annotated with the custom YOLOv5m model.
        }
    \label{fig:ann}
\end{figure}

The Mask-RCNN was effective in accurately annotating the apples in the generated training data. However, one limitation of this method was that the Mask-RCNN network also annotated the apples located on the ground. This became problematic while testing the trained YOLOv5 models on the MinneApple dataset, since the apples on the ground are not annotated in the entire MinneApple dataset. To overcome this challenge, it was essential to ensure that the YOLOv5 model, which was trained on the generated dataset, could distinguish between apples present on the ground and those in the trees. Therefore, when utilizing the auto-annotation technique, it was necessary to ensure that the apples on the ground were not annotated. Figure \ref{fig:ann} presents an example, where the apples on the ground are annotated.

In order to tackle this problem, the MinneApple training set was utilized to train a YOLOv5m model \cite{yolov5} for automatically annotating apples present in trees. To generate annotations, a customized annotation pipeline was employed, which utilized the YOLOv5 model trained on this dataset to predict the positions of apples in the dataset. As the MinneApple dataset solely contains annotated apples on apple trees, this automatic annotation technique is able to accurately identify only those apples that are located on the trees and not on the ground, as shown in Figure \ref{fig:ann}. Using this automatic annotation technique, the annotations for the entire generated dataset were produced. 

In addition, this new annotation technique offers the advantage of being able to predict the location of more apples in the generated images compared to the method that utilizes Mask-RCNN, as illustrated in Figure \ref{fig:ann}.

It should be taken into consideration that when the annotation process of a dataset is aided by a detector trained on the MinneApple train set, the resulting detector trained on generated data cannot outperform the baseline. This is due to the fact that any apples that the baseline fails to predict will not be included as annotations in the generated dataset, preventing the network from learning about these deficiencies.

\subsection{Detection Method}
Using YOLOv5 \cite{yolov5}, a pipeline was created to train a YoloV5m model on the generated Stable Diffusion images. The YOLOv5m model was preferred over YOLOv5s because it has a larger architecture and the model can operate on low-end hardware that is suitable for mounting on UAVs, unlike YOLOv5l and YOLOv5x, which have mostly higher memory requirements. Opting for the m-model instead of the s-model is justified by another factor: the YOLOv5m model demonstrates enhanced precision in predicting smaller objects compared to YOLOv5s \cite{small_object}.

The images are preprocessed by resizing them to 1280x704 pixels, which is close to the size of the MinneApple dataset \cite{apple_database} while still being usable in the YOLOv5 pipeline.

Within the pipeline to train a YoloV5m model, the Adam optimizer \cite{adam} was utilized, incorporating a learning rate of $10^{-4}$. The training process spanned 50 epochs with a batch size of 8. In order to evaluate the performance of the models, the MinneApple test data was employed during the testing phase.

\subsection{Evaluation}

Evaluation is a critical component of training machine learning models. It helps to understand how well models perform and how it compares to other models. This section will discuss evaluation metrics and model comparisons.

\subsubsection{Evaluation Metrics}
\label{metrics}

The Intersection over Union (IoU) and Average Precision (AP) are both commonly used in evaluating the performance of object detection algorithms. In the context of AP calculation, IoU is used to determine whether a predicted bounding box overlaps sufficiently with the ground truth bounding box to be considered a true positive. Specifically, IoU is calculated by dividing the area of overlap between the predicted bounding box $B_p$. and the ground truth bounding box $B_{gt}$ over the area of their union $\mathnormal{B_p} \cup \mathnormal{B_{gt}}$, see Equation \ref{eq: IOU}. 

\begin{equation}
    \label{eq: IOU}
    \mathnormal{a_o} = \frac{\mathnormal{B_p} \cap \mathnormal{B_{gt}}}{\mathnormal{B_p} \cup \mathnormal{B_{gt}}} 
\end{equation}

If the IoU is greater than a predefined threshold, the predicted bounding box is considered a true positive. Once the true positives are identified using the IoU threshold, the average precision (AP) is calculated by computing the precision at a fixed set of equally spaced recall levels and taking the mean. 

The definition of average precision follows as used in the Pascal VOC Challenge \cite{pascalvoc}, as can be seen in Equation \ref{eq:ap}. Where $R$ is a set of evenly spaced recall levels ($r$), for research, $\{ 0.0, 0.01 ..., 1.0\} $ as used by MS-COCO \cite{mscoco}. Furthermore, $p_{interp}(r)$ is the interpolated maximum precision at a given recall level. The AP value can be used to evaluate the overall performance of the object detection model at different IoU thresholds.

\begin{align}
    \label{eq:ap}
    \text{AP} := \frac{1}{|\mathcal{R}|}\sum_{r \in \mathcal{R}} p_{interp}(r), && p_{interp}(r) := \max_{\tilde{r}: \tilde{r} \geq r} p(\tilde{r})
\end{align}

In this study, the average precision will be measured at three distinct thresholds, corresponding with the evaluation methodology used for the MinneApple dataset \cite{apple_database}. These thresholds are AP@50, AP@75, and the threshold range of AP@0.5:0.05:0.95.

\subsubsection{The Baseline}

In order to evaluate the effectiveness of training the YOLOv5m model on synthesized Stable Diffusion image data, a baseline model will be trained on the annotated MinneApple dataset. The goal of this comparison is to determine how the synthesized data-trained model performs compared to the real-world data-trained model, in terms of average precision, defined as Equation \ref{eq:ap}. To ensure that the comparison is fair and unbiased, the two models will be tested on the same dataset, the MinneApple dataset test data. This will provide important insights into the efficiency of training object detection models with synthesized Stable Diffusion image data.

\section{Experiments, Results \& Discussion}

\begin{figure}[H]
    \centering
    \includegraphics[width=0.455\textwidth, height=0.12\textheight]{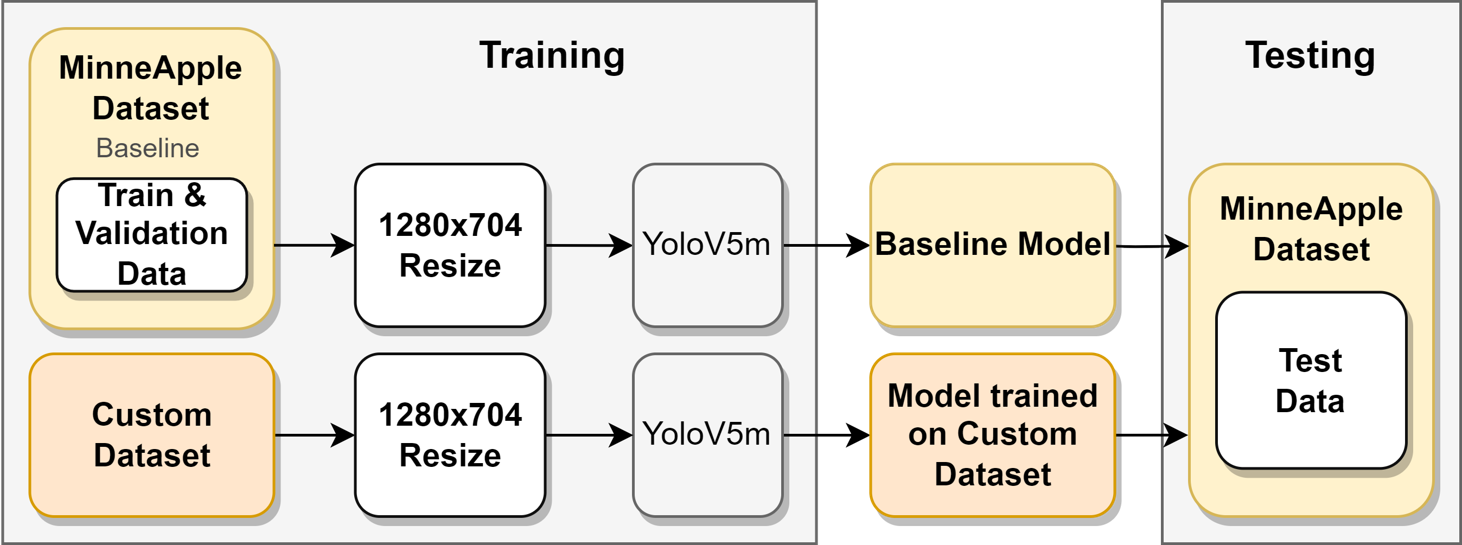}
    \caption{
        Experimental Setup}
    \label{fig:ex}
\end{figure}

The experiments in this part of the paper involved training YOLOv5m models on both the MinneApple training set and the generated dataset. After the models were trained, they were evaluated using the MinneApple testing data. Figure \ref{fig:ex} provides a visual representation of the experimental setup. To assess the performance of the models, several evaluation metrics were calculated according to the criteria described in subsection \ref{metrics}. Once the experiments were five times repeated, the mean of the average AP score for each model is determined, along with the standard deviation of these scores. The results obtained from these experiments provide insights into the performance of the YOLOv5m models in detecting apples, and how well they perform on generated datasets compared with real-world data.

\subsection{Metrics}

The evaluation metrics for two datasets, MinneApple as a baseline and the generated dataset, along with their respective differences, are presented in Table \ref{tab:eval}. Both datasets were used to train a YOLOv5m model, and the metrics reported in the table were obtained through the evaluation of the trained models on the MinneApple test data. The table shows the \textit{AP} for different \textit{IoU} thresholds, namely \textit{AP@0.50}, \textit{AP@0.5:0.05:0.95}, and \textit{AP@0.75}. The reported values for each metric are the mean values along with their corresponding standard deviations (SD) obtained over five runs of the experiments.

For the baseline, the model achieves the highest performance in terms of AP@0.50 with a score of $0.70$, which indicates that the model can detect objects with an IoU of at least 50\%. The model also performs reasonably well in terms of AP@0.5:0.05:0.95 and AP@0.75, with scores of $0.36$ and $0.34$, respectively.

For the generated dataset, the model achieves a slightly lower performance compared to MinneApple. The model achieves an AP@0.50 score of $0.61$, indicating that the model is able to detect the objects in the images with an IoU of at least 50\%. The scores for AP@0.5:0.05:0.95 and AP@0.75 are $0.30$ and $0.25$, respectively. The difference between the generated dataset compared to the baseline is $\textbf{0.09}$, $\textbf{0.06}$, and $\textbf{0.09}$ for AP@0.50, AP@0.5:0.05:0.95, and AP@0.75, respectively.

These results suggest that the model trained on the generated dataset is able to detect the objects in the images with moderate accuracy.

\begin{table}[htbp]
    \centering
    \caption{Evaluation metrics}
    \begin{tabular}{ c || c  c  c } \toprule
    
        \textbf{Dataset} & \textbf{AP@0.50} & \textbf{AP@0.5:0.05:0.95} & \textbf{AP@0.75}\\ \midrule

        Baseline & 
            \textbf{0.70} \small{± 0.008} & 
            \textbf{0.36} \small{± 0.011} &
            \textbf{0.34} \small{± 0.022} \\ 
         
        Generated & 
            0.61 \small{± 0.013} & 
            0.30 \small{± 0.010} & 
            0.25 \small{± 0.016} \\ \midrule \midrule

        \textit{Difference} & 
            $\textit{0.09}$ & 
            $\textit{0.06}$ & 
            $\textit{0.09}$ \\ \bottomrule
            
    \end{tabular} 

    \begin{minipage}[t]{1\linewidth}
	\end{minipage}

    \label{tab:eval}
\end{table}

\subsection{Apple Detection}
\begin{figure}[H]
    \centering
    \includegraphics[width=0.45\textwidth, height=0.3\textheight]{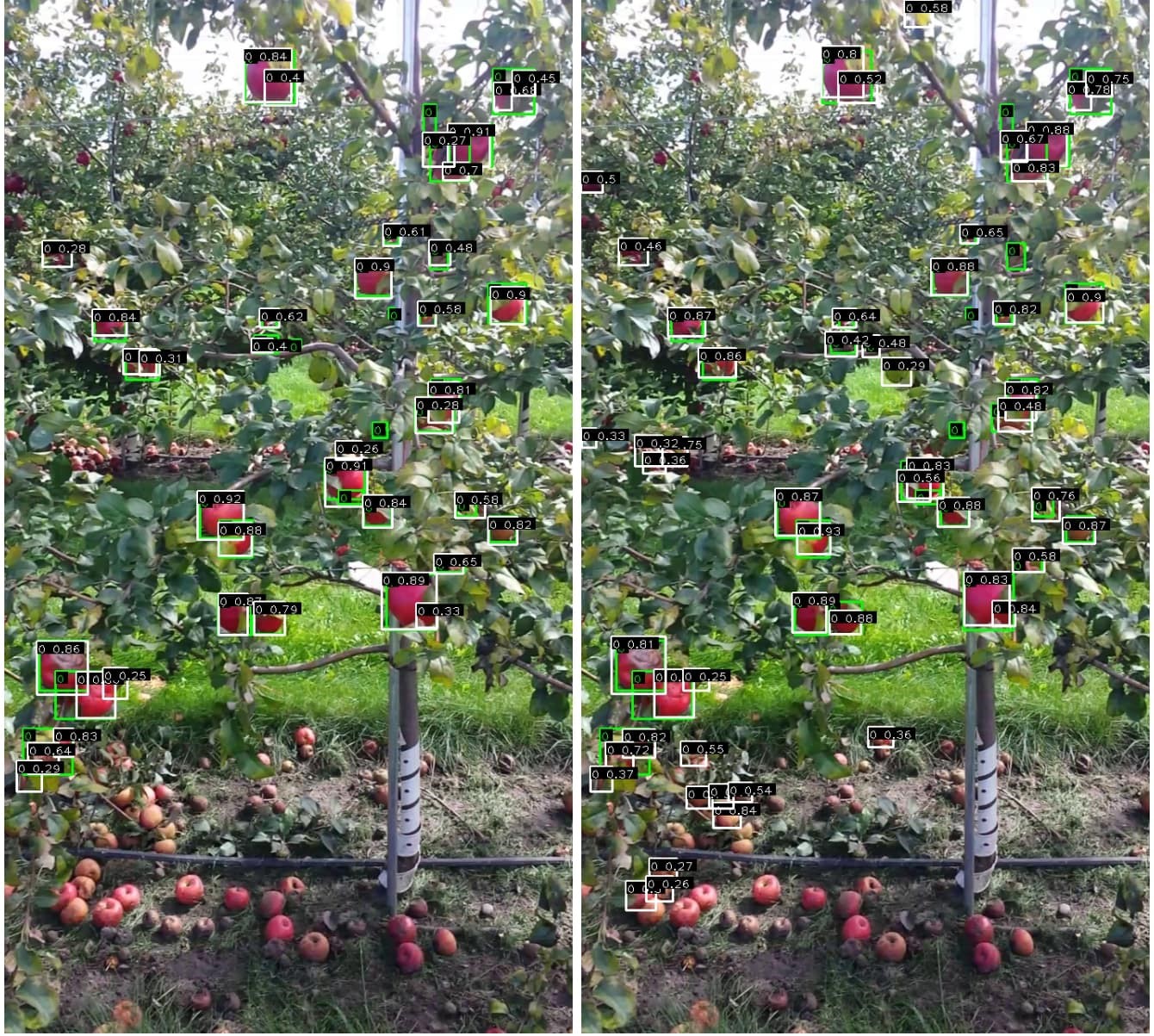}
    \caption{
        A sample image of the MinneApple test dataset, where the apples are predicted by YOLOv5m models that were trained on both the MinneApple dataset (\textit{left}) and the generated dataset (\textit{right}) are depicted by white bounding boxes. The ground truth is indicated by the green bounding boxes. See Appendix \ref{app:predictions} for additional examples.
        }
    \label{fig:detect_1}
\end{figure}

A more in-depth analysis of the differences between the two YOLOv5m models trained on distinct datasets can be achieved by examining their qualitative results. In Figure \ref{fig:detect_1}, a sample image from the MinneApple test set is shown. The green bounding boxes in both images represent the ground truth, which was provided by the MinneApple test set. In the \textit{left} image, the white bounding boxes represent the predictions made by the YOLOv5m model trained on the MinneApple training data, which serves as the baseline. Meanwhile, the white bounding boxes in the \textit{right} image are the predictions made by the YOLOv5m model trained on the generated dataset. These results demonstrate an impression of the performance of the models on the detection task, where both models are able to detect the apples in the trees well.

In terms of qualitative differences, the model trained on the generated dataset stands out. In a few occasions, it still detects apples on the ground even when the annotations of these apples in the generated data are not present in the dataset. Specifically, the model predicts apples on the ground only when there are foliage or leaves surrounding or near the apples, as illustrated in Figure \ref{fig:detect_1}. This observation may support the hypothesis that the model trained on the generated dataset is learning to identify and locate apples in the presence of nearby foliage, as there are no instances of apples in combination with leaves on the ground in this dataset. Since the apples on the ground are not annotated but predicted, these predictions may affect the performance of the model in comparison with the baseline.

\begin{figure}[H]
    \centering
    \includegraphics[width=0.45\textwidth, height=0.17\textheight]{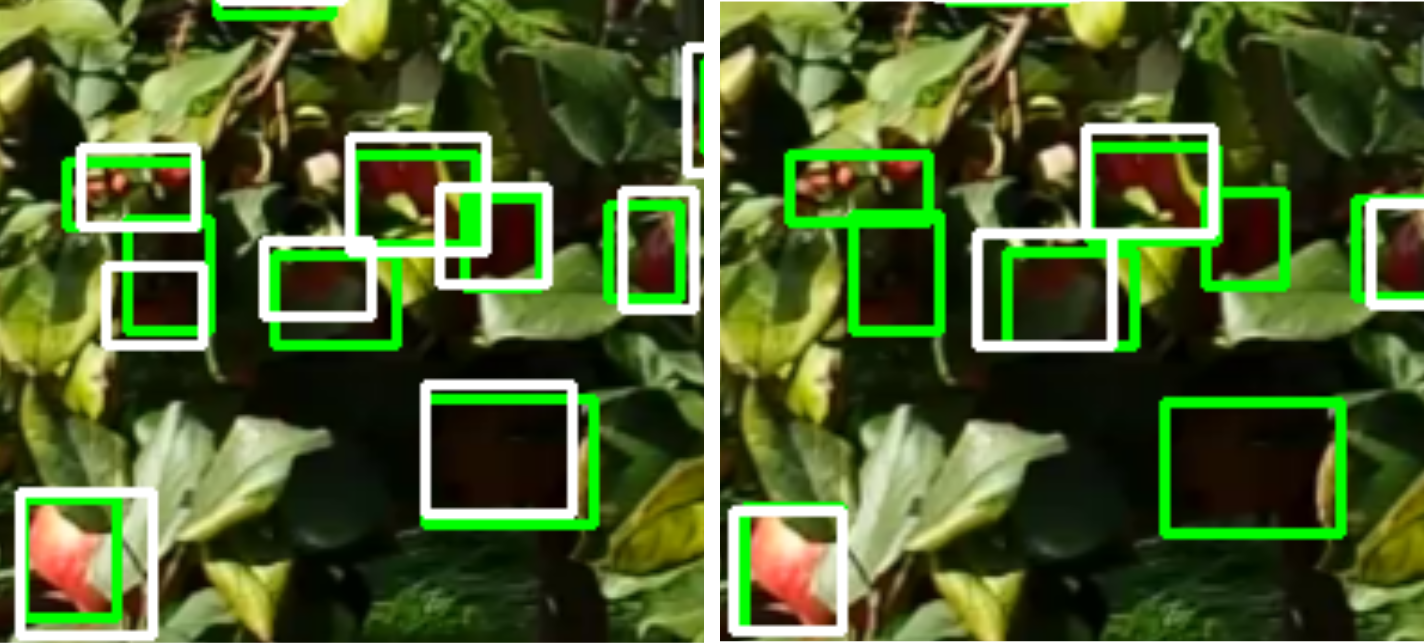}
    \caption{
        A Cropped sample image of the MinneApple test dataset, where the apples are predicted by YOLOv5m models that were trained on both the MinneApple dataset (\textit{left}) and the generated dataset (\textit{right}) are depicted by white bounding boxes. The ground truth is indicated by the green bounding boxes.
        }
    \label{fig:detect_shadows}
\end{figure}

The YOLOv5m model trained on generated data has another minor limitation, in that it has difficulties detecting apples in areas that are heavily occluded located in regions with high levels of shading. It frequently experiences difficulty identifying apples in such conditions, whereas the baseline model is able to predict the location of these apples. However, the apples that are not detected by the model trained on generated data, are also barely visible to the naked eye due to these heavy shadows, as can be seen in Figure \ref{fig:detect_shadows}. As the model trained on the generated dataset fails to predict these apples, this limitation may affect its performance and account for the observed differences compared to the baseline model.

For further discussions, as previously stated, the model trained on the generated dataset is unable to outperform the baseline model, potentially due to the use of the MinneApple dataset to generate annotations for the new dataset.

Considering all of these findings, these results emphasize the potential of this approach to improve the current data collection process in the realm of computer vision, with only a slight variance in performance.

\section{Conclusion}
The proposed study has shown the potential of using artificially synthesized images to enhance the current data collection process. In order to address the research question "\textit{How might Latent Diffusion be utilized to synthesize images of apple orchard, that resemble real-world like scenarios?}", it has been suggested that Latent Diffusion models, such as Stable Diffusion, have the potential to be employed in our application for generating apple tree images, when using highly detailed prompt crafted trough prompt engineering. These images can subsequently be utilized to train deep-learning algorithms. To simulate apple orchards, the Stable Diffusion model was given Positive- and Negative-prompts as stated in section \ref{sec: iamge synthesis}.

Although the synthesized images are realistic, it should be noted that the generated images are not flawless. When attempting to recreate apple trees to be compared to the MinneApple dataset, the generated images possess a deficiency in heavy shadows in the apple tree and foliage surrounding the apples on the ground. This ultimately has consequences on the performance of the trained model on these images, particularly in terms of average precision, when compared to a baseline model trained on the original MinneApple dataset.

Furthermore, to answer the question “\textit{What is the performance of a deep learning model trained on synthetic data versus a deep learning model trained on real data?}”, the results of this study indicate that deep learning models, for this research YOLOv5m, trained on synthetic data, slightly underperforms to those trained on real data in terms of qualitative performance, as stated in the aforementioned limitations.

Moreover, the difference between the average precision of the generated dataset and the baseline model is $\textbf{0.09}$, $\textbf{0.06}$, and $\textbf{0.09}$ for AP@0.50, AP@0.5:0.05:0.95, and AP@0.75, respectively. Although underperforming, these findings emphasize the potential of this automated approach to offer a more efficient and cost-effective data collection method across various industries.

The results of this study indicate that the hypothesis stating that deep learning models trained on synthetic data will exhibit comparable performance to models trained on real data in real-world scenarios is not fully supported. The findings demonstrate that the models trained on real data consistently outperformed those trained on synthetic data, indicating a performance gap between the two.

However, despite the underperformance of the models trained on synthetic data, the study highlights the potential of synthetic data as a viable alternative or supplement to real data in deep learning model training. Although the models trained on synthetic data did not achieve comparable performance, they still showed promising results and demonstrated the capacity to generalize to some extent in real-world scenarios.

Moving forward, it is crucial to continue exploring the potential use of Latent Diffusion models for synthesizing images that can be utilized for data collection and training deep learning models. This exploration can open up new opportunities for practical applications and expedite progress in various fields.

\section{Future work}

To begin, expanding the scope of this image dataset synthesis approach to other detection tasks beyond apples could be a valuable avenue for future research. It could offer insights into the generalizability of the approach and its potential applications in other domains. Going beyond the realms of object detection and computer vision in a broader sense, this study has demonstrated that Stable Diffusion can generate remarkably realistic images. The potential use cases extend to domains such as film, animation, advertisement, marketing, and virtual prototyping.

Although this study sheds light on the performance of models trained on generated datasets using prompts compared to a baseline, there are opportunities for further investigation. One potential avenue for future research is to enhance the shading in the synthesized images using Prompt Engineering, which would result in more realistic images. Currently, models trained on the generated dataset lack the ability to predict the location of apples in areas with heavy shading. In this study, efforts have already been made to create images with more shadows, as evidenced by Appendix \ref{app:apples_shading}, but more extensive research is required to increase the shadows in the images.

Furthermore, the use of prompt engineering should be further explored to generate images that feature both foliage and apples on the ground. The present images only depict apples on the ground and lack this characteristic. Despite nearby foliage, the final model still predicts the apples on the ground.

It would be compelling to investigate hybrid approaches that blend Prompt Engineering with actual photographs of apple trees. The generated data could supplement the real-life data to create a larger dataset, with the aim of enhancing the model's performance. While this approach has been investigated for addressing class imbalance in datasets \cite{dataset_imbalance}, it would be intriguing to observe its impact when there is no class imbalance but rather a scarcity of real-world data.

To augment the automatic data collection process, it would be beneficial to explore automatic annotations. Although the present study employs a model trained on a baseline dataset to automatically annotate apples, it would be desirable to have a method that does not necessitate initial data to establish a model that can annotate synthesized images. The recently developed state-of-the-art technique, Segment Anything Model (SAM) \cite{sam}, could possibly be utilized for this purpose.

\section{Broader Impact}

The utilization of Stable Diffusion in this study for image dataset synthesis highlights an important consideration. It is critical to recognize that, while this technique provides significant benefits, both Stable Diffusion and other generative models can pose ethical challenges.

The use of synthesized images raises issues of data generation, bias, and authenticity. Diverse and representative training datasets are essential for avoiding bias development. Failure to capture real-world scenarios can result in object detectors that are biased and discriminatory. 

Furthermore, there is a risk of deceptive use of synthesized images, undermining digital media integrity and trust. Misuse of this technology could lead to the creation of convincing forgeries, deep fakes, or misleading visual information, which could have far-reaching consequences for individuals and society. 

However, image synthesis, utilizing Stable Diffusion, enables the creation of realistic and diverse images. Particularly in object detector training, this may lead to improved object detection algorithms. This progress contributes to enhanced safety and efficiency across various domains, including autonomous vehicles, surveillance systems, and medical imaging. Furthermore, the synthesis of large-scale datasets enables training on rare or hazardous scenarios, ultimately enhancing performance in real-world applications.

In summary, while image synthesis using Stable Diffusion holds tremendous potential for improving object detection algorithms, it is vital to recognize the ethical considerations associated with data generation potential misuse. By proactively addressing these concerns, the power of this technology can be utilized responsibly, to ensure its broader impact aligns with ethical principles.

\acknowledgements{
	\noindent This project was financially supported by Regieorgaan SIA (part of NWO) and performed within the RAAK PRO project Mars4Earth. We would like to thank our collaborators at Saxion University of Applied Sciences for insightful discussions.
}

\begin{figure}[H]
    \begin{minipage}[c]{0.4\linewidth}
        \includegraphics[width=\linewidth]{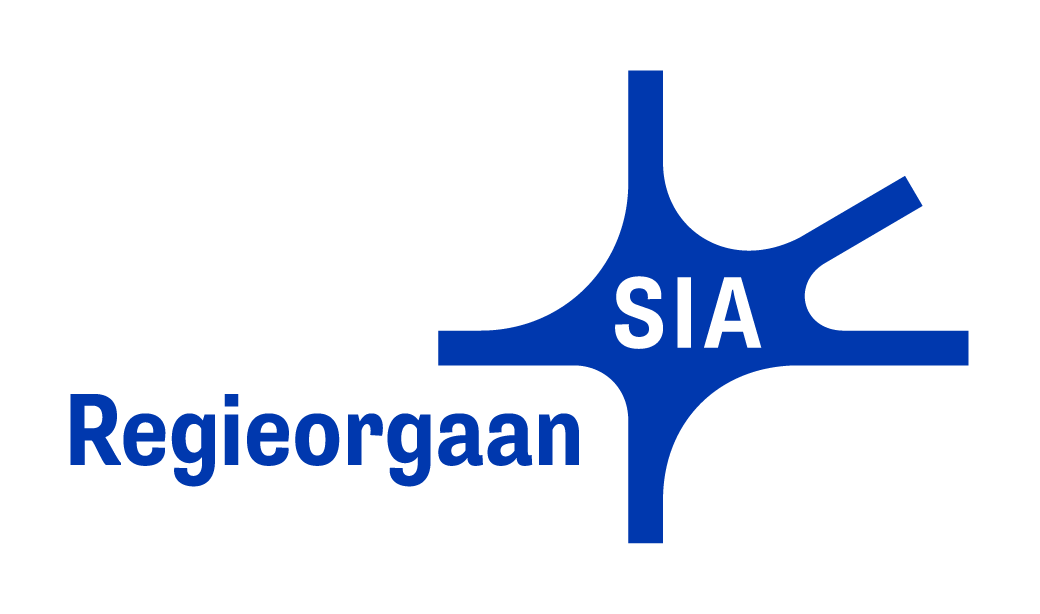}
    \end{minipage}
    \hfill
    \begin{minipage}[c]{0.4\linewidth}
        \includegraphics[width=\linewidth]{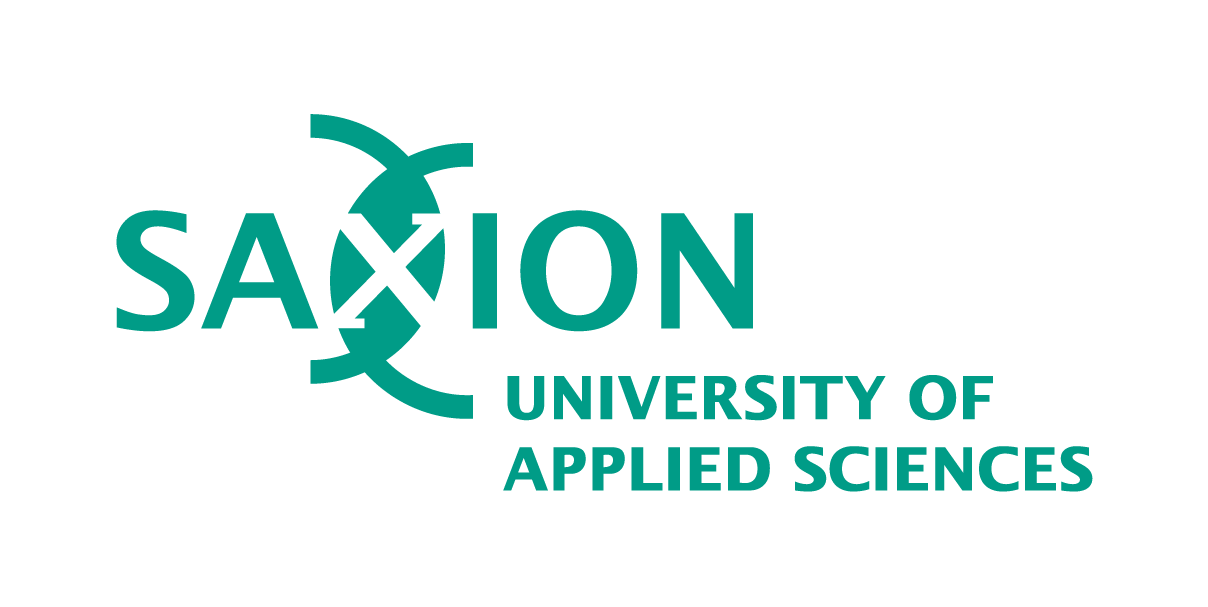}
    \end{minipage}%
\end{figure}

\bibliographystyle{unsrt}
\bibliography{Exploring_the_Effectiveness_of_Dataset_Synthesis}

\clearpage
\begin{appendices}
    \onecolumn
    \section{Synthesized Images}
    \label{app:synthesize-images}

    \vfill
    \begin{figure}[h]
        \centering
        \includegraphics[width=1\textwidth, height=0.6\textheight]{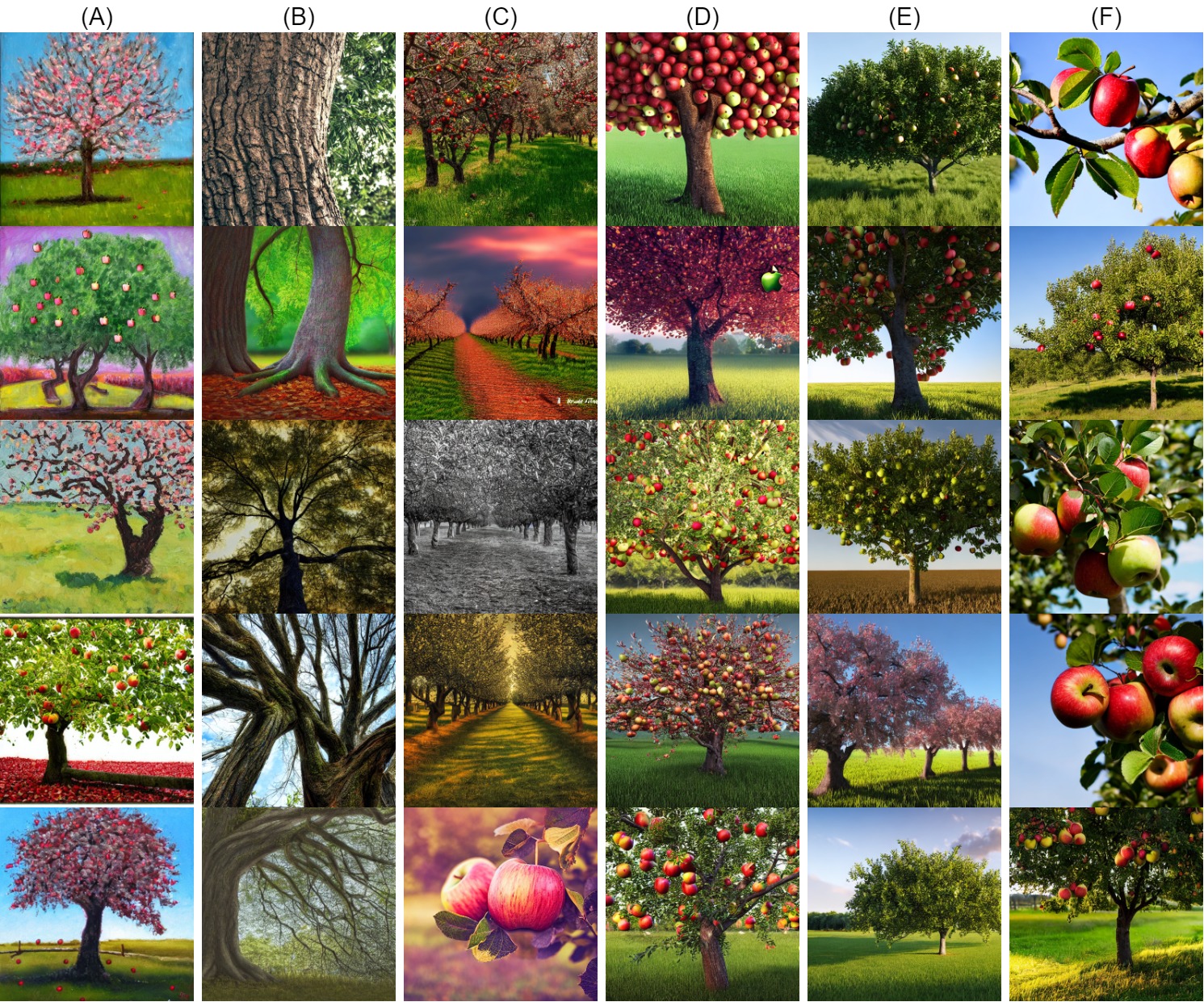}
        \caption{Synthesized images using Stable Diffusion, with the default model configurations. A) Prompt: \textit{"Apple trees"}. B) Prompt: \textit{"photo of a tree, hyper-realism, 4k, realistic, photograph"}. C) Prompt: \textit{"apple orchard, hyper-realism, 4k, realistic, photograph"}. D) Prompt: \textit{"apple tree with many apples, apples, hyper-realism, 4k, render, cinematic lighting"}. E) Positive prompt: \textit{prompt D)}; Negative prompt: \textit{"blurry image, deformed, cartoon, drawing"}. F) Positive prompt: \textit{"photo of a tree branch with apples, apple tree, many apples, hyper-realism, 4k, realistic, photograph"}; Negative prompt: \textit{negative prompt E)}.}
        \label{fig:apple_1}
    \end{figure}
    \vfill
    
    \begin{figure}[p]
        \centering
        \includegraphics[width=1\textwidth, height=0.6\textheight]{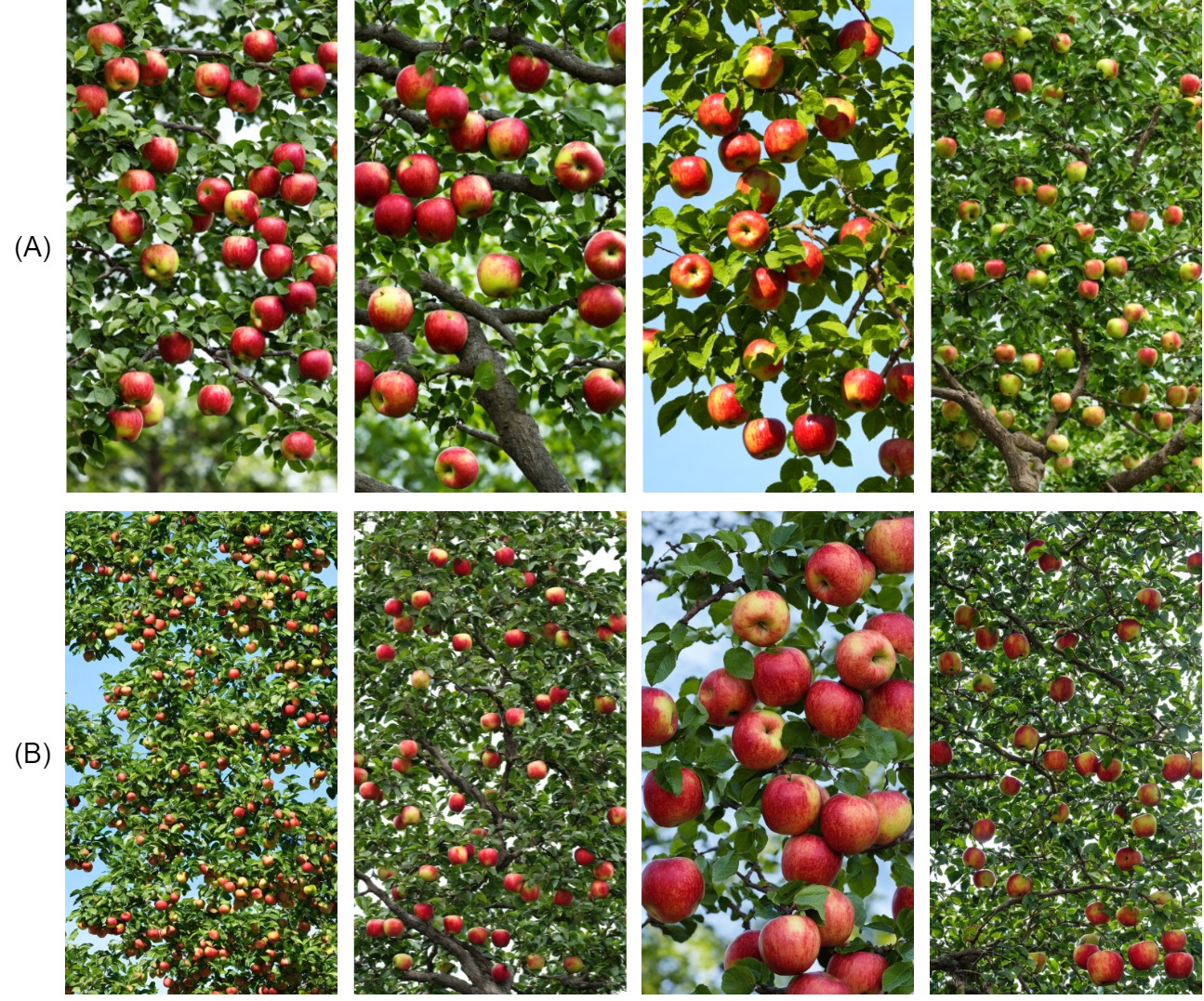}
        \caption{Synthesized images using Stable Diffusion, with the default model configurations. A) Positive prompt: \textit{"photo of a tree branch with apples, apple tree, many apples, hyper-realism, 4k, realistic, photograph"}, Negative prompt: \textit{"blurry image, deformed, cartoon, drawing"}, image size: 1280x704. B) Prompt: \textit{positive prompt \& negative prompt: prompts A)}, image size: 1280x704, CFG: 6, inference steps: 30.}
        \label{fig:apple_2}
    \end{figure}

    \begin{figure}[p]
        \centering
        \includegraphics[width=1\textwidth, height=0.6\textheight]{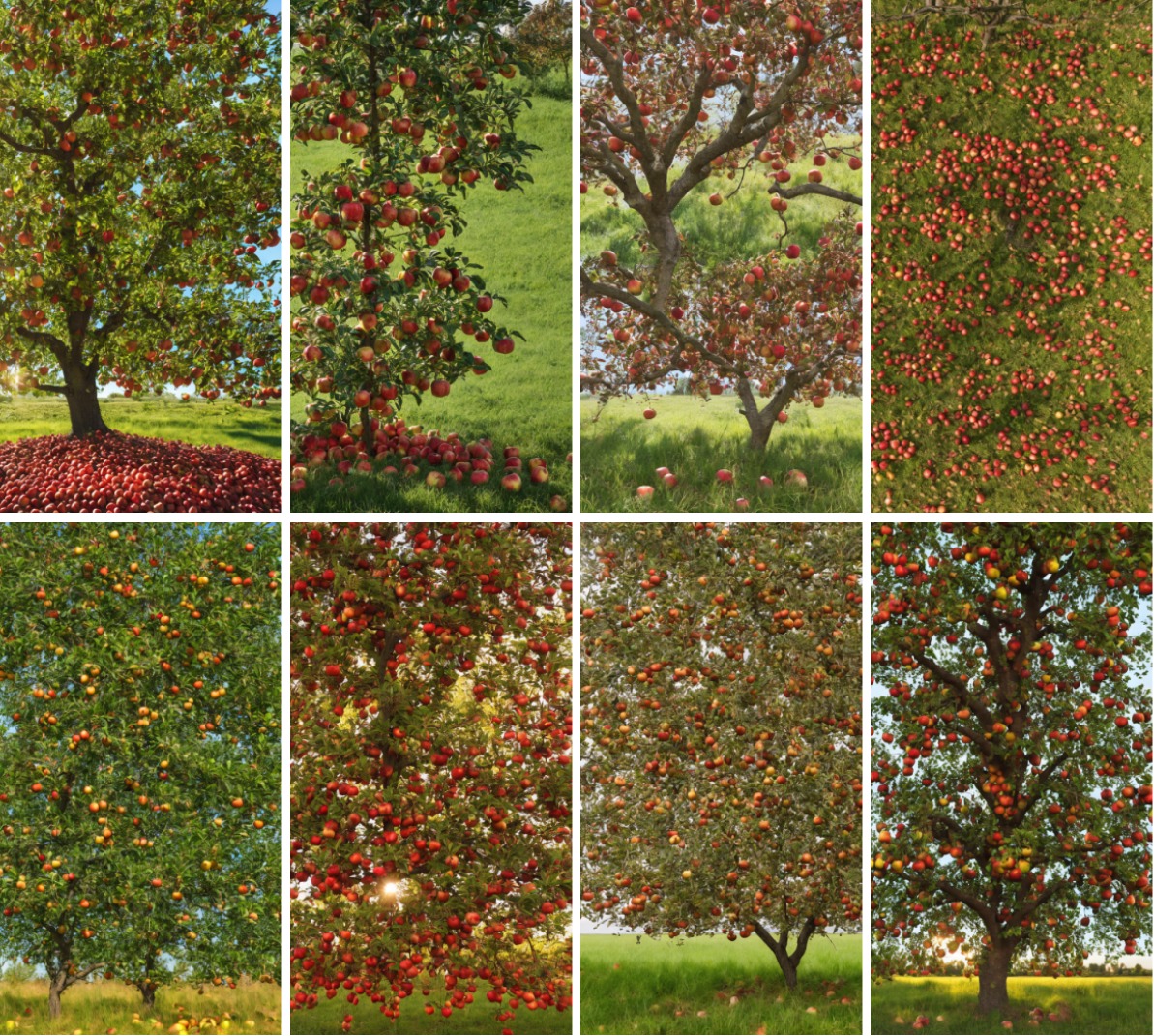}
        \caption{Synthesized images using Stable Diffusion, with the default model configurations. Positive prompt: \textit{"a photo of a tree standing in the grass. the tree has many apples, the apples are both red and yellow. beneath the tree there are a lot of apples. The many apples are a combination of red apples and yellow apples. volumetric lighting. shadows, hyperrealism, 4k realism, photograph"}, Negative prompt: \textit{"blurry image, deformed, cartoon, drawing, painting"}, image size: 1280x704. CFG: 6, inference steps: 30.}
        \label{fig:apple_final_big}
    \end{figure}
    
    \clearpage
    \section{Apple Prediction Images}
    \label{app:predictions}

    \vfill
    \begin{figure}[h]
        \centering
  	\hspace{0.01\textwidth}
	  \begin{minipage}{1\textwidth}
	       \centering
              \subfloat[\textbf{Baseline Model} \label{fig:pr_b}]{\includegraphics[width=1\textwidth]{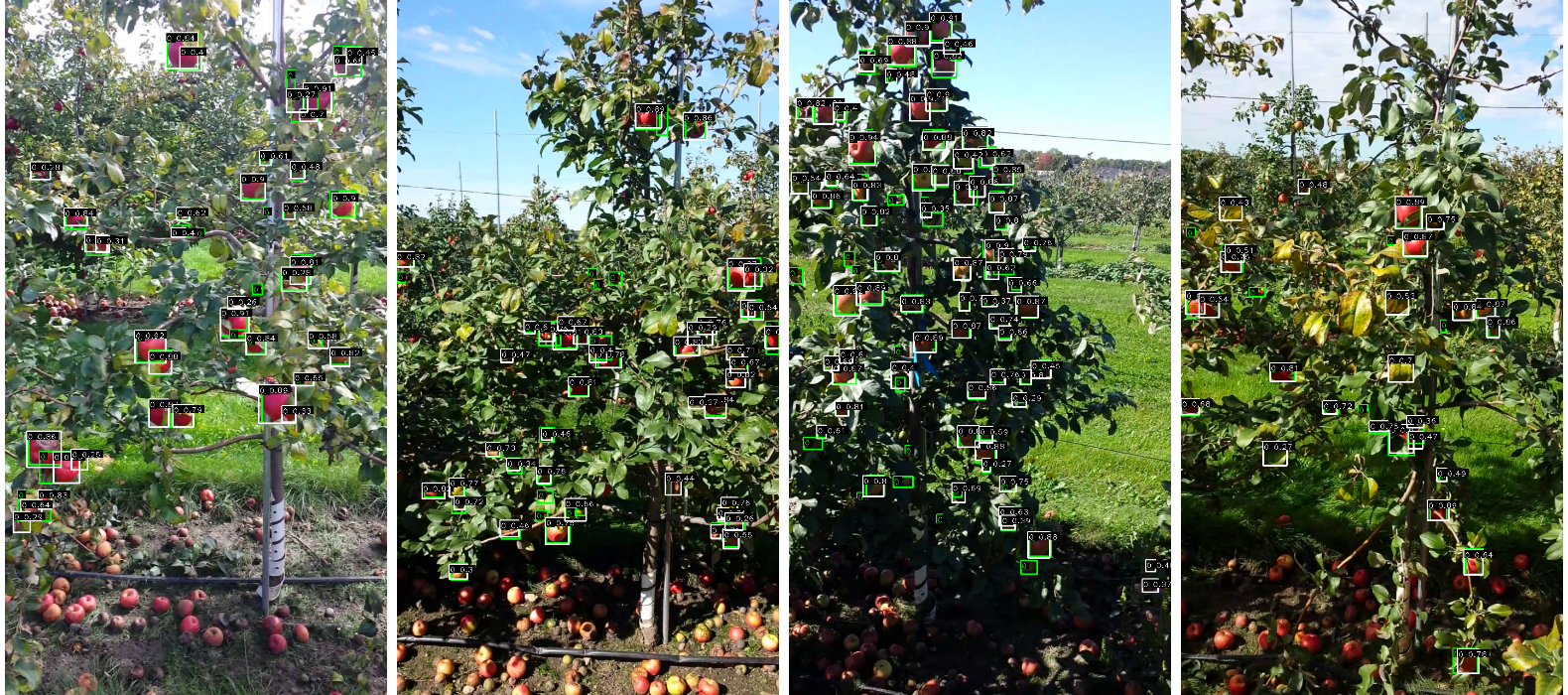}}
	  \end{minipage}
	\hspace{0.01\textwidth}
 	\begin{minipage}{1\textwidth}
	       \centering
              \subfloat[\textbf{Model Trained on Generated Dataset} \label{fig:pr_g}]{\includegraphics[width=1\textwidth]{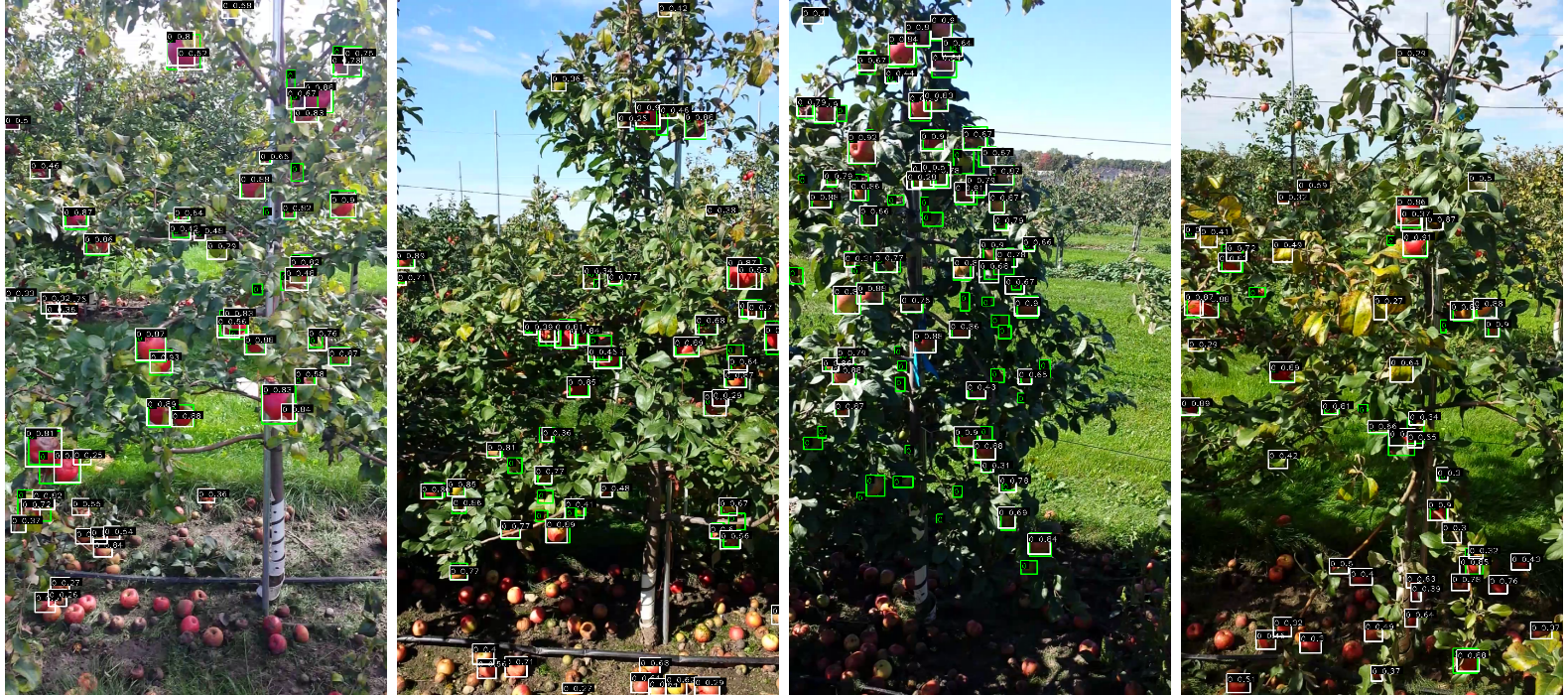}}
	  \end{minipage}
    \caption{Sample images of the MinneApple test dataset, where the apples are predicted by YOLOv5m models that were trained on both the MinneApple dataset (\textit{a}) and the generated dataset (\textit{b}) are depicted by white bounding boxes. The ground truth is indicated by the green bounding boxes.}
    \label{fig:qualitative-results}
    \end{figure}
    \vfill

    \clearpage
    \section{Future Work: Appels in the Shading}
    \label{app:apples_shading}

    \vfill
    \begin{figure}[h]
        \centering
        \includegraphics[width=0.65\textwidth, height=0.65\textheight]{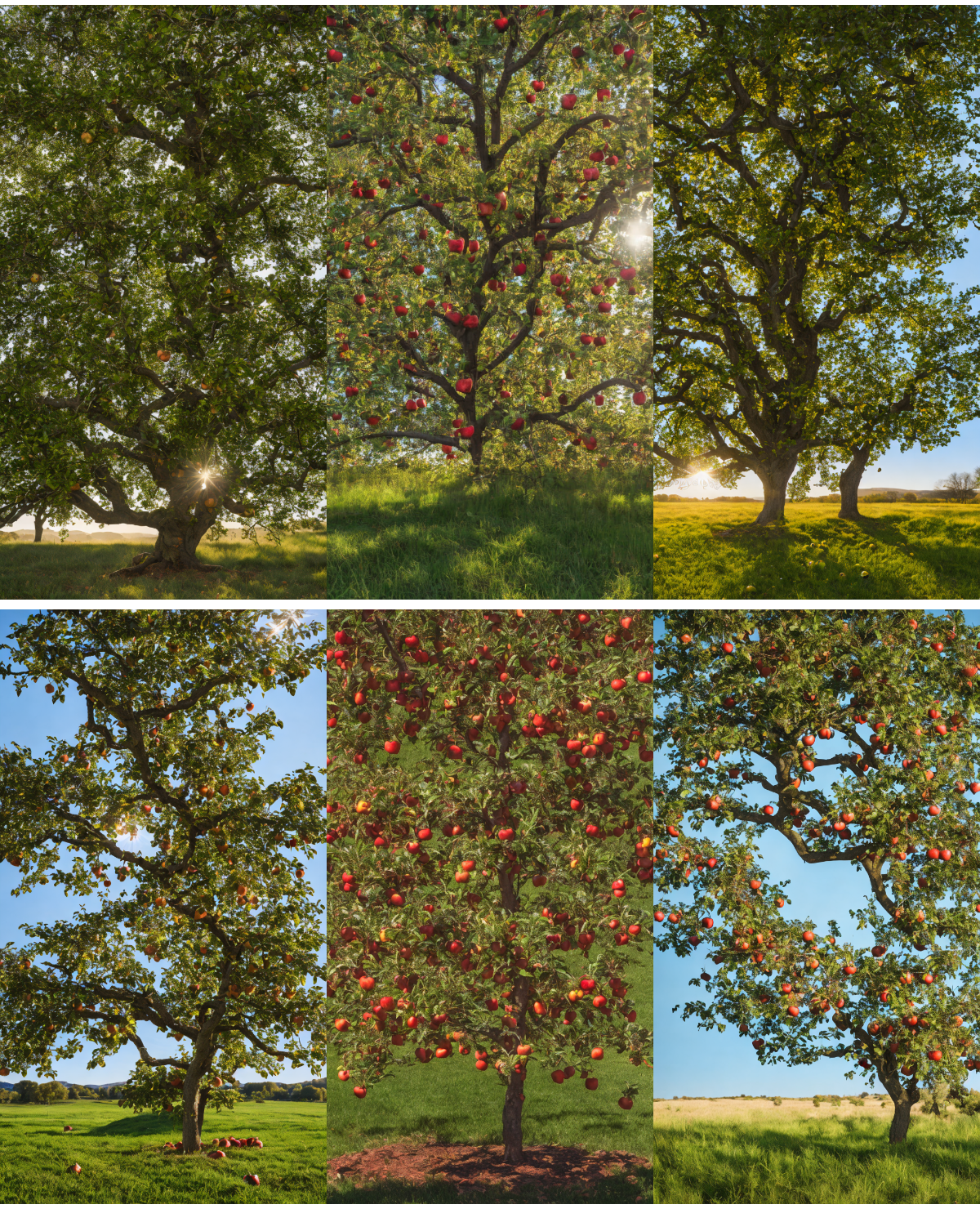}
        \caption{Synthesized images with an attempt to include shadows, using Stable Diffusion, with the default model configurations. Positive prompt: \textit{"a photo of a tree standing in the grass the, tree is partly in the shadow. the tree has many apples in the tree that are both red (apples) and yellow (apples). beneath the tree there are a lot of apples. cinematic lighting, lots of fine details, hyper-realistic, real shadow, dark setting, ultra photorealistic dramatic shadows"}, Negative prompt: \textit{"blurry, deformed, cartoon, drawing, treeless, painting"}, image size: 1280x704, CFG: 6, inference steps: 30.}
        
        \label{fig:shadow_big}
    \end{figure}
    \vfill

\end{appendices}
\end{document}